\documentclass[10pt, a4paper, twocolumn, teaser, showabstract]{naverlabseurope1}

\usepackage{lipsum}
\usepackage{multicol}
\usepackage{tikz,tkz-kiviat,pgfplots}
\usepackage{tabularx}
\usepackage{booktabs}
\usepackage{amssymb}
\usepackage{amsmath}
\usepackage{float}
\usepackage[utf8]{inputenc}

\usepackage{microtype}
\usepackage{graphicx}
\usepackage{natbib}

\graphicspath{{figures/}}

\title{Retrieval-augmented generation in multilingual settings}

\correspondingauthor{[nadia.chirkova,vassilina.nikoulina]@naverlabs.com}

\authors{
Nadezhda Chirkova
 \quad David Rau \quad Hervé Déjean\quad \\ \textbf{Thibault Formal}\quad  \textbf{Stéphane Clinchant}\quad \textbf{Vassilina Nikoulina}
}
\affiliations{NAVER LABS Europe}
\contributions{}
\website{https://github.com/naver/bergen}
\websiteref{\href{https://github.com/naver/bergen}}

\teaserfig{\includegraphics[width=0.75\textwidth]{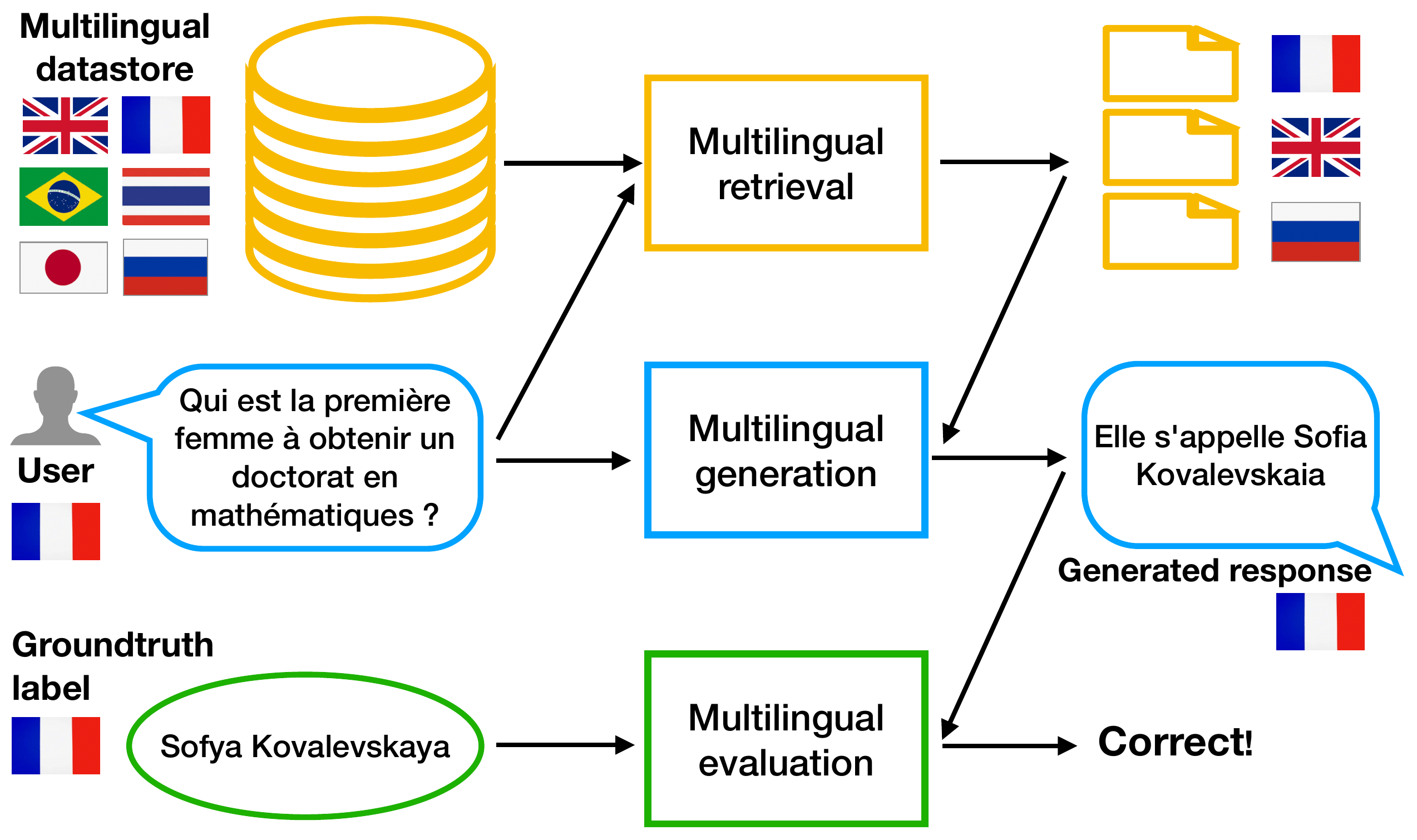}}
\teasercaption{\label{fig:ill} Multilingual retrieval-augmented generation pipeline. We study which components are required to build a well performing mRAG pipeline, that can be used as a strong baseline in future works.}

\begin{abstract}

Retrieval-augmented generation (RAG) has recently emerged as a promising solution for incorporating up-to-date or domain-specific knowledge into large language models (LLMs) and improving LLM factuality, but is predominantly studied in English-only settings. In this work, we consider RAG in the multilingual setting (mRAG), i.e. with user queries and the datastore in 13 languages, and investigate which components and with which adjustments are needed to build a well-performing mRAG pipeline, that can be used as a strong baseline in future works. Our findings highlight that despite the availability of high-quality off-the-shelf multilingual retrievers and generators, task-specific prompt engineering is needed to enable generation in user languages. Moreover, current evaluation metrics need adjustments for multilingual setting, to account for variations in spelling named entities.  The main limitations to be addressed in future works 
 include frequent code-switching in non-Latin alphabet languages, occasional fluency errors, wrong reading of the provided documents, or irrelevant retrieval.
 We release the code for the resulting mRAG baseline pipeline at \url{https://github.com/naver/bergen}$^1$.
\end{abstract}

\begin{document}

\maketitle

\definecolor{codegreen}{RGB}{93, 168, 128} % Light pleasant blue
\definecolor{codeblue}{RGB}{74, 74, 250}  % Pleasant light green
\definecolor{codered}{RGB}{209, 106, 114}
\newcommand{\vn}[1]{{\textcolor{red}{VN:#1}}}

\newcounter{mycounter}
\newcommand\modelcounter{\stepcounter{mycounter}\textbf{\themycounter}}
\newcommand\labelledmodelcounter[1]{\hspace{-0.5cm}\refstepcounter{mycounter}\textbf{(\themycounter)}\label{model:#1}}
\newcommand{\modelref}[1]{\textbf{(\ref{model:#1})}}

\newcommand\blfootnote[1]{%
  \begingroup
  \renewcommand\thefootnote{}\footnote{#1}%
  \addtocounter{footnote}{-1}%
  \endgroup
}

\section{Introduction}
Retrieval-augmented generation (RAG)~\citep[\textit{inter alia}]{lewis_retrieval-augmented_2020,ram-etal-2023-context} has recently emerged as a promising solution for incorporating up-to-date or domain-specific knowledge into large language models (LLMs) and improving LLM factuality, especially in knowledge-intensive tasks such as open-domain question answering or fact-checking.
% \begin{figure}[t!]
%     \centering
%     \includegraphics[width=0.9\linewidth]{figs/illustration.pdf}
%     \caption{Multilingual retrieval-augmented generation pipeline. We study which components are required to build a well performing mRAG pipeline, that can be used as a strong baseline in future works.}
%     \label{fig:ill}
% \end{figure}
\begin{table*}[!h]
     \centering
    \begin{tabular}{c|c|cccc} %|p{0.3cm}p{0.3cm}p{0.3cm}p{0.3cm}}
    \toprule
    %& \multicolumn{4}{c|}{MKQA} & \multicolumn{4}{c}{XOR-TyDi QA} \\ 
      & No  & \multicolumn{4}{c}{Retrieval from Wiki in} \\
     & retrieval & English & User lang & English+UL & All langs \\
     \midrule
     MKQA & & & &  & \\ \midrule
     English & 58.4 & \textbf{70.2} &--- & --- &68.5 \\ 
     Arabic & 26.4 & 45.9 & 36.3 & \textbf{49.0} & 48.2\\
     Chinese & 21.4 & 29.1 & 22.5 & 27.2 & \textbf{31.0} \\
     French & 48.4 & 62.6 & 56.3 & 65.0 & \textbf{66.2}\\ 
     Finnish$^\ddagger$ &  29.7 & 55.8 & 45.2 & 
    59.8 & \textbf{60.7}\\ 
     German & 47.8 & 64.6 & 54.8 & 65.5 & \textbf{66.9}\\
     Italian &  51.5 & 61.2 & 56.8 & 64.8 & \textbf{66.3}\\
     Japanese & 31.7 & \textbf{42.7} & 28.8 & 40.2 
    & 42.1\\
     Korean &21.5 & 32.2 & 31.5 & \textbf{38.4}  & 38.1\\
     Portuguese & 48.4 & 62.3 & 54.9 & 65.2 & \textbf{66.9}\\
     Russian$^\dagger$ &38.1 & 55.0 & 51.0 & \textbf{61.0} & 59.4\\
     Spanish & 52.5 & 63.3 & 57.3 & 65.7 &  \textbf{67.1} \\
     Thai$^\ddagger$ &   12.4 & 23.7 & 10.1  & 23.2 & \textbf{24.5} \\
     \midrule
     XOR TyDi QA & & & &  &\\ \midrule
     English & 47.5 &  \textbf{64.2} & ---& --- & 59.4\\
     Arabic & 47.7 & 52.9 & 65.5 & 66.6 & \textbf{66.8} \\
     Finnish$^\ddagger$ & 30.8 & 45.2 & 58.9 & \textbf{60.9} & 59.1\\
     Japanese & 21.0 & 25.2 & 30.0 & 24.8 & \textbf{31.8} \\
     Korean & 31.0 & 33.4 & 40.8 & 40.0 & \textbf{41.8}\\
     Russian$^\dagger$ & 40.5 & 53.9 & 62.3 & 63.8 & \textbf{64.6}\\
     %& ko & fr & ru & en & ko & fi & ru & ar \\
    %No retrieval & 18.6 & 52.6 & 36.2 & 55.6 & 29.5 &  & 39.4 &  \\
    %Retrieval in En & 33.9 & 66.5 & 54.9 & 75.3 & 34.3 &  & 53.6 & \\ 
    %Retrieval in UL & 29.6 & 55.1 & 49.4 & --- & 40.3 &  & 63.9  & \\
    \bottomrule
\end{tabular}
    \caption{
    Performance of mRAG for various languages on MKQA and XOR-TyDi QA datasets (TyDi QA for English), with different retrieval options. Metric: character 3-gram recall. Retriever: BGE-m3. Reranker: BGE-m3. 
    %(both include all considered languages in training).
    Generator: Command-R-35B. Prompt: translated into user languages with an instruction to generate in the given user language (UL). $^\dagger$ denotes languages included in Command-R pretraining but not instruction tuning. $^\ddagger$ denotes languages not included in Command-R pretraining nor  tuning. %; \newline
    \textit{RAG brings substantial performance improvement in all languages, and retrieval from multilingual Wikipedia is beneficial in most cases.}
    }
    \label{tab:1}
\end{table*}
RAG augments user queries with relevant context retrieved from the Internet or a given collection and then passes the result to an LLM to generate a knowledge-grounded response. 
Recent works focus on improving various components of the complex RAG pipeline,
%RECENTLINK, 
e.g. generator~\citep{yoran2024making}
%retrieval collection~\citep{tamber_wiki} 
or search query processor~\citep{ma-etal-2023-query}, as well as addressing fragility of the RAG approach, e.g. filtering irrelevant retrieved context~\citep{wang2023learning,xu2023recomp,kim2024sure} or dynamically deciding for which user queries retrieval is actually needed~\citep{jiang-etal-2023-active,asai_self-rag_2023}.

Unfortunately, all listed efforts are focusing on English as the data language in their experiments, i.e. the language of the user queries and of the knowledge datastore. In this work, we argue for the importance of considering multilingual settings in RAG experiments and advancing multilingual RAG (mRAG), as it has clear advantages for both English and non-English speakers. On the one side, enabling access to RAG advances for non-English speakers requires testing the applicability of approaches proposed in the literature for non-English queries, and possibly developing special multilinguality-oriented RAG methodologies. On the other side, considering non-English knowledge datastores ensures access to local or culture-specific information for all future users of RAG models, as such information is often available only in non-English. In the similar way retrieving from English may be beneficial for non-English queries e.g. about US or British culture.

Enabling high-quality RAG in multilingual settings requires access to strong multilingual retrievers and generators, as well as high-quality multilingual evaluation. The retriever should be able to map queries in the user language to the documents in the same or different language. The generator should be able to generate fluently and correctly in the user language, but also to understand documents in various languages and to follow instructions specified in the prompt. %, e.g. \textit{"Answer to this question as short as possible, relying on the provided support documents."} 
While recent advances in natural language processing and information retrieval made appropriate candidate components available, the entire multilingual RAG pipeline was not evaluated in the literature before.

The \textit{main contribution} of our work is (1) building a publicly available baseline mRAG pipeline, to foster research on multilingual RAG in a zero-shot setting, and (2) conducting an initial study of mRAG in open question answering with user queries and retrieval datastores in 13 languages. We build on top of BERGEN, a benchmarking library for RAG~\citep{bergen}. We aim to answer the following research questions:
\begin{itemize}
    \item does RAG bring same performance improvements in knowledge-intensive tasks in non-English as in English?
    \item which components are needed for effective mRAG and which adaptations are required?
    \item what are the main limitations of the existing components that can be addressed in future work?
\end{itemize}

Our key findings can be summarized as follows:
\begin{itemize}%[noitemsep,topsep=0pt,parsep=0pt,partopsep=0pt]
    \item Retrieval: recent off-the-shelf multilingual retrievers and rerankers perform reasonably well in both cases when queries and documents are in the same or different language, and also handle well retrieval from multilingual datastores 
    (Tables \ref{tab:1} and \ref{tab:retrieval});
    \item Generation: achieving high performance across all languages requires a strong multilingually pretrained and tuned LLM, coupled with advanced prompting, e.g. translating prompts into user languages and instructing the LLM to generate responses in the user language (Tables~\ref{tab:prompts_demo}, \ref{tab:prompts} and \ref{tab:models});
    \item Evaluation: evaluation metrics need adjustment to take into account the zero-shot scenario, e.g. variations in spelling named entities in cross-lingual settings (Table~\ref{tab:charrecall});
    \item The main limitations to be addressed in future works 
    include frequent code-switching\footnote{Code-switching refers to inserting fragments in other languages when generating in a given language.} in non Latin alphabet languages, occasional fluency errors, wrong reading of the provided documents, or irrelevant retrieval (Table~\ref{tab:inspection}).
\end{itemize}

\section{Related Work}
Despite mRAG being not well studied in the literature, some of  the individual components of the RAG pipeline were rather well developed for multilingual settings, e.g. multilingual retrievers and generator LLMs; we discuss them in Section~\ref{sec:main}.

The closest line of work to ours is multilingual open question answering \citep[\textit{inter alia}]{cora_asai,muller-etal-2022-cross,sorokin-etal-2022-ask,asai-etal-2022-mia} defined as a the task of answering non-English questions from a large collection of multilingual documents, as introduced in~\citep{cora_asai}. 
Those aforementioned works train task-specific models combining cross-lingual retrievers and multilingual generation models, e.g. with iterative extension of annotated data used in the CORA approach~\citep{cora_asai}. The key difference of our work is that we compose the mRAG system in a \textit{zero-shot manner}, using off-the-shelf components without dedicated training. This  approach, dominating nowadays in the literature, is enabled by recent advances in  LLMs and retrieval and makes the system more robust and easy-to-extend. 
It's important to note that our goal is not to outperform the mentioned models such as CORA, but to evaluate the state of the described zero-shot mRAG setting, understand its open problems, and  provide an experimental ground for future development of mRAG.

\section{Multilingual RAG pipeline}
\label{sec:main}

The high-level illustration of the mRAG pipeline is presented in Figure~\ref{fig:ill}. The input is represented by a \textit{user query} $q$ in language $L_q$. This could be an arbitrary user request to an LLM. Following the common practice of testing RAG systems on open-domain question answering, we assume $q$ is an information-seeking question. The model is expected to output response $r$ which correctly answers the given question. An important (and reasonable) expectation is that the model replies in the user language, i.e. $r$ is written in $L_q$. 

\begin{table}[]
    \centering
    \footnotesize
    \begin{tabular}{p{2.2cm}p{4cm}p{0.5cm}} 
        \toprule
        \textbf{Prompt label} & \textbf{Prompt text (written in the language specified in the last column)}  & \textbf{Prom. lang.} \\
        \midrule
        \verb|Reply short (EN)| & \textit{``Answer a given question as short as possible.''} & EN \\ \midrule
        \verb|Reply short in| \verb|same lang (EN)| & \textit{``Answer a given question as short as possible.  Answer in the same language as the language of the question.''} & EN \\ \midrule
        \verb|Reply short in UL| \verb|(EN)| & \textit{``Answer a given question as short as possible.  Answer in \{UL\}.''} & EN \\ \midrule
        \verb|Reply short (UL)| & \textit{``Answer a given question as short as possible.''} & UL \\ \midrule
        \verb|Reply short in UL| \verb|(UL)| & \textit{``Answer a given question as short as possible. Answer in \{UL\}.''} & UL \\ \midrule
        \verb|Reply short in UL| \verb|+ NE in UL (UL)| & \textit{``Answer a given question as short as possible. Answer in \{UL\} and write all named entities in \{UL\} alphabet.''} & UL \\
        \bottomrule
    \end{tabular}
    \caption{System prompts used in our experiments. \{\textit{UL}\} denotes a placeholder to insert the target language.
    }
    \label{tab:prompts_demo}
\end{table}

\begin{table}[]
    \centering
    \footnotesize
    \begin{tabular}{p{1.05cm}p{2.5cm}p{3.2cm}} 
        \toprule
         & \textbf{Text}  & \textbf{Character 3-grams} \\
        \midrule
         Ground truth & sofya kovalevskaya & [\underline{sof} ofy fya \underline{kov} \underline{ova} \underline{val} \underline{ale} \underline{lev} \underline{evs} \underline{vsk} \underline{ska} kay aya] \\
         Model response & sofia kovalevskaia & [\underline{sof} ofi fia \underline{kov} \underline{ova} \underline{val} \underline{ale} \underline{lev} \underline{evs} \underline{vsk} \underline{ska} kai aia] \\ \midrule
         Recall & 0 & $9/13=69.2\%$ \\
        \bottomrule
    \end{tabular}
    \caption{Illustration of the proposed character 3-gram recall metric, designed to be more robust to  different possible transliterations of named entities. Tokens matching between groundtruth and model response are underlined. 
    }
    \label{tab:charrecall}
\end{table}

\textbf{Step 1: retrieval.} 
The first step in mRAG is retrieving \textit{context} $c$ relevant to the query $q$ from the Internet or a particular \textit{collection} $C$, using the \textit{retriever system} $R$: $c = R(\tilde q, C), \tilde q = Q(q)$. Here $Q$ denotes an optional query generation model which infers a search query $\tilde q$ from a user query $c$, e.g. it can be an LLM prompted to reformulate the query, or simply copying the user query $q$. Following a standard practice in testing RAG systems, we use Wikipedia as our collection $C$. In most of the experiments we assume monolingual $C$ in language $L_C$ (English or user language), but we also experiment with retrieving from the multilingual $C$. 

The retriever system $R$ usually consists of two stages. The first stage \textit{ranker} $R_1$ encodes queries $q$ and documents $d\in C$ independently: $h_q = R_1(\tilde q) \in \mathbb{R}^{n}$, $h_d = R_1(d)  \in \mathbb{R}^{n}$, allowing to precompute document representations offline and enabling fast search over large collections, e.g. $ \tilde c = \texttt{top-K}_{d \in C} h_q^T h_d$, $K$ denotes the number of retrieved documents. The second-stage \textit{reranker} $R_2$ processes a (small) subset $\tilde c$ of documents from $C$ retrieved by $R_1$ and encodes documents together with queries: $h_{q, d} = R_2(\tilde q, d) \in \mathbb{R}$, enabling semantically richer representations and selecting $k$ most relevant documents: $c = \texttt{top-k}_{d \in \tilde c} h_{q, d}$. Both $R_1$ and $R_2$ are often based on BERT-like models and trained on  retrieval datasets such as MS-MARCO~\citep{nguyen2016ms}. % multilingual and cross-lingual
In our work we rely on retrievers and rerankers developed specifically for the multilingual setting. 

\textbf{Step 2: generation.} 
The second stage of mRAG pipeline consists of generating a response $r$ based on the user query $q$ and retrieved relevant context $c$ with a generator LLM: $r = \mathbb{LLM}(q, c)$. State-of-the-art LLMs follow the wide-spread paradigm of pretraining a decoder-only Transformer model on a large set of unsupervised data and then tuning it for instruction following and alignment with user preferences. This second step of instruction tuning and alignment often introduces a \textit{template}, representing formatting rules for passing data into the LLM. Template usually contains placeholders for user queries $q$, model responses $r$ and also for a \textit{system prompt}, which is put in the beginning of the template and describes the task / role for the LLM. A simplest example of the system prompt is \textit{``You are a helpful assistant.''}. In our work we study several generator LLMs and experiment extensively with various prompting strategies for mRAG.

Below we describe how we instantiate different components of our mRAG pipeline.

\paragraph{Multilingual retrievers.} 
The described problem setting requires strong monolingual and cross-lingual rankers and rerankers, for cases when $L_q = L_C$ and $L_q \ne L_C$, correspondingly. We pick a strong recently released and publicly available \verb|BGE-m3|\footnote{Retriever: \url{https://huggingface.co/BAAI/bge-m3} (dense version). Reranker: \url{https://huggingface.co/BAAI/bge-reranker-v2-m3}.}~\citep{chen2024bge} which provides all listed functionalities and includes all languages we consider in its training data.
%and proprietary \verb|Cohere.Embed| (???) for our experiments. 
We also consider a baseline including query translation, where query generator $Q$ translates $q$ from $L_q$ to $L_C$. 
% specify particular models snd links, translation model
We employ the NLLB-600M translation model\footnote{\url{https://huggingface.co/facebook/nllb-200-distilled-600M}} \citep{nllbteam2022language}.

\paragraph{Multilingual generation.} 
Most of current state-of-the-art LLMs are either English-centric or support a limited set of languages, possibly due to under-investigated effects of the "curse of multilinguality" for large models~\citep{conneau-etal-2020-unsupervised}, i.e. it is yet unclear how many languages LLMs can fit without hurting performance, or due to limited availability of multilingual instruction tuning and alignment datasets. At the same time, it was shown that even English-centric LLMs, which were pretrained and finetuned mostly on English data, may exhibit good multilingual capabilities due to the occasional presence of multilingual data in pretraining~\citep{ye2023language,chirkova2024}. As such, we experiment with both strong English-centric and recent multilingual models. Among English-centric models we pick commonly-used \verb|LLaMA-2-7B-chat|~\citep{touvron2023llama} and state-of-the-art \verb|SOLAR-10.7B|~\citep{kim2023solar}, and among multilingual models we pick \verb|Mixtral-8x7B|~\citep{jiang2024mixtral} and \verb|Command-R-35B|\footnote{\url{https://huggingface.co/CohereForAI/c4ai-command-r-v01}}. All models were instruction-tuned. \verb|Command-R-35B| was  developed with keeping RAG application in mind and officially supports 11 languages\footnote{Command-R official languages: Arabic, Brazilian Portuguese, English, French, German, Italian, Japanese, Korean, Simplified Chinese, and Spanish}, including most of our considered languages, and also includes 13 more languages (incl. Russian) in pretraining but not instruction tuning. \verb|Mixtral-8x7B| was pretrained on the multilingual data with 5 languages\footnote{Mixtral official languages: English, French, Italian, German, and Spanish}, we use it's instruction-tuned version. 

\paragraph{System prompt.}  In our preliminary experiments we noticed that models sometimes reply in English even for non-English user queries. This is not an expected behavior and substantially reduces metrics, calculated over groundtruth answers in user languages. To tackle this, we study various  strategies for defining the system prompt, e.g. including an explicit  instruction to reply in the user language, see Table~\ref{tab:prompts_demo} for all the system prompts that we consider. Some strategies include translation of the prompts into user languages: we used Google Translate and asked native or fluent speakers of considered languages, employed in our research laboratory, to check and correct the generated translations\footnote{Issues raised when controlling prompt translation include (1) wrong semantics of the assistant's task in translations which is highly undesirable; (2) choosing between formal and informal register -- we chose informal style for all cases; (3) complications with translating field-specific terms such as ``named entities''; (4) absence of the direct translation of the phrase "You are a helpful assistant" in some languages.}.

\begin{table}[]
    \centering
    \scriptsize
    \begin{tabular}{p{1.3cm}p{0.05cm}p{0.05cm}p{0.05cm}p{0.05cm}p{0.05cm}p{0.05cm}p{0.05cm}p{0.05cm}p{0.05cm}p{0.05cm}p{0.05cm}p{0.05cm}p{0.05cm}}
\toprule
MKQA & en & ar & es & fi & fr & de & ja & it & ko & pt & ru & th & zh \\
\midrule
\# examples & \multicolumn{13}{c}{2827}\\
len ques. & 43 & 38 & 48 & 46 & 49 & 47 & 26 & 48 & 22 & 45 & 42 & 41 & 16 \\
len answ. & 11 & 10 & 11 & 11 & 11 & 11 & 8 & 11 & 6 & 11 & 10 & 12 & 6 \\
\bottomrule
\end{tabular}

\vspace{0.1cm}

\begin{tabular}{lp{0.1cm}}
\toprule
Tydi QA & en  \\
\midrule
\# examples & 440  \\
len ques. & 39  \\
len answ. & 13 \\
\bottomrule
\end{tabular} ~~~
\begin{tabular}{lp{0.1cm}p{0.1cm}p{0.1cm}p{0.1cm}p{0.1cm}}
\toprule
XOR-Tydi QA & ar & fi & ja & ko & ru  \\
\midrule
\# examples & 708 & 615 & 433 & 371 & 568  \\
len ques. &  30 & 37 & 18 & 20 & 42 \\
len answ. & 11 & 14 & 5 & 5 & 11  \\
\bottomrule
\end{tabular}

\vspace{0.1cm}

\begin{tabular}{p{1.3cm}p{0.05cm}p{0.05cm}p{0.05cm}p{0.05cm}p{0.05cm}p{0.05cm}p{0.05cm}p{0.05cm}p{0.05cm}p{0.05cm}p{0.05cm}p{0.05cm}p{0.05cm}}
\toprule
Wikipedia & en & ar & es & fi & fr & de & ja & it & ko & pt & ru & th & zh \\
\midrule
\# ex. (M) & 25 & 3.3 & 10 & 1.5 & 13 & 14 & 27 & 8.2 & 1.6 & 4.7 & 8.6 & 3.7 & 11  \\
len pass. & 624  & 585 & 619 & 833 & 627 & 720 & 208 & 650 & 431 & 619 & 721 & 217 & 206  \\
\bottomrule
\end{tabular}
\caption{Statistics of the used data. Len denotes median length in Unicode characters.
}
\label{tab:stats}
\end{table}

\paragraph{Multilingual QA datasets.} 
We follow~\citet{cora_asai} and use MKQA~\citep{longpre-etal-2021-mkqa} and XOR-TyDi QA~\citep{asai-etal-2021-xor} datasets for evaluation in our experiments. MKQA consists of 10k examples from the Natural Questions (NQ) dataset~\citep{kwiatkowski2019natural}, translated into 25 languages. This dataset is therefore parallel between languages and grounds knowledge primarily in English Wikipedia. In our experiments we select a subset of 2.7K samples, overlapping between MKQA and KILT NQ datasets\footnote{NQ dataset in KILT benchmark available at \url{https://huggingface.co/datasets/kilt_tasks}}, thus recovering relevant documents information from KILT NQ. XOR-TyDi QA comprises 40K information-seeking questions in 7 languages (of which we us 3K validation questions) and grounds questions in Wikipedia in the same language as the question or in English. To provide English for comparison, we include results for English on the TyDi QA dataset~\citep{clark-etal-2020-tydi}. Though both datasets come with oracle contexts, questions are context-independent, meaning that they can be understood without context and the answers are ``universal'' and not specific to the provided contexts. This property is not held for many other multilingual QA datasets, e.g. some reading comprehension datasets.

Statistics of the used datasets (number of examples, average lengths) are presented in Table~\ref{tab:stats}. We select a diverse set of user languages (ULs) to experiment with, including Latin and non Latin script ones (see Table~\ref{tab:1}).

\begin{table*}[t!]
\centering
\footnotesize
\begin{tabular}{p{3.8cm}|lll|lll|lll|lll}
\toprule
& \multicolumn{6}{c|}{Correct language rate (CRL)} & \multicolumn{6}{c}{Character 3-gram recall} \\ \midrule
& \multicolumn{3}{c|}{SOLAR-10.7B} & \multicolumn{3}{c|}{Command-R-35B} &   \multicolumn{3}{c|}{SOLAR-10.7B} & \multicolumn{3}{c}{Command-R-35B} \\ \midrule
 & ko & fr & ru &  ko & fr & ru  & ko & fr & ru  & ko & fr & ru  \\
\midrule 
 \multicolumn{13}{c}{No retrieval} \\ 
 \midrule
%\labelledmodelcounter{prompt_en_nor} 
Reply short (EN) & 7.6 & 47.3 & 50.7 &  94.2 & 85.1 & 88.5 &  12.1 & 50.1 & 26.9  & 22.6 & 49.0 & 33.5  \\
~+ reply in same lang (EN) & 25.7 & 70.8 & 69.1 &  91.8 & 84.3 & 84.9 & 10.5 & 47.0 & 27.4 &  21.9 & 47.1 & 31.9   \\
~+ reply in UL (EN) & 60.5 & 94.1 & 84.7 &  99.2 & 92.0 & 93.7 & 11.0 & 48.0 & 31.1 & 21.9 & 49.2 & 32.2   \\
%\labelledmodelcounter{prompt_ul_nor} 
Reply short (UL) & 1.0 & 73.6 & 46.3 &  99.8 & 92.1 & 95.3 & 12.6 & 52.8 & 27.1 &  22.9 & 49.4 & 35.4   \\
~+ reply in UL (UL) & 51.5 & 97.3 & 97.5 &  99.9 & 92.0 & 98.1 &  11.2 & 51.0 & 33.8 &  21.9 & 47.7 & 36.4   \\
~+ reply in UL + NE in UL (UL) & 61.9 & 98.6 & 98.2 &  99.6 & 97.4 & 98.3 &  11.2 & 50.8 & 33.6 &  18.6 & 52.6 & 36.2   \\ \midrule
 \multicolumn{13}{c}{Retrieval in English} \\ \midrule
%\labelledmodelcounter{prompt_en_enr} 
Reply short (EN) & 21.1 & 71.8 & 61.0 &  54.3 & 47.2 & 41.7  & 17.3 & 64.1 & 41.3 &  23.8 & 59.8 & 32.5  \\
~+ reply in same lang (EN) & 51.9 & 91.2 & 90.9 & 67.8 & 64.3 & 53.5 & 17.7 & 64.3 & 52.5 & 24.8 & 60.6 & 35.0   \\
~+ reply in UL (EN) & 83.4 & 99.4 & 98.1 & 96.8 & 89.6 & 80.6 &  19.5 & 64.1 & 55.6 & 29.8 & 60.4 & 41.7   \\
%\labelledmodelcounter{prompt_ul_enr} 
Reply short (UL) & 2.8 & 90.1 & 59.4 & 98.3 & 96.8 & 94.7 & 17.9 & 64.4 & 41.4 & 30.0 & 62.6 & 50.1   \\
~+ reply in UL (UL) & 69.3 & 99.5 & 99.5 &100 & 98.6 & 96.5 &  18.6 & 64.6 & 56.6 &  33.7 & 62.8 & 53.2   \\
~+ reply in UL + NE in UL (UL) & 53.1 & 99.7 & 99.7 & 100 & 99.5 & 97.8 &  18.4 & 64.5 & 56.7 & 33.9 & 66.5 & 54.9  \\  \midrule
 \multicolumn{13}{c}{Retrieval in user languages} \\ \midrule
%\labelledmodelcounter{prompt_en_ulr} 
Reply short (EN) & 24.7 & 76.9 & 70.0 &  99.9 & 95.8 & 97.4 &  16.0 & 55.8 & 44.6 & 28.4 & 51.7 & 46.9  \\
~+ reply in same lang (EN) & 32.3 & 92.0 & 91.0 & 99.9 & 96.8 & 97.5 & 18.0 & 55.5 & 49.4 & 28.7 & 51.3 & 46.6  \\
~+ reply in UL (EN) & 61.9 & 99.4 & 95.8 & 100 & 97.3 & 97.5 & 22.2 & 55.9 & 50.4 & 28.8 & 51.5 & 46.5  \\
%\labelledmodelcounter{prompt_ul_ulr} 
Reply short (UL) & 9.0 & 90.3 & 78.4 & 100 & 98.9 & 98.9 & 15.4 & 55.7 & 47.1 &  29.0 & 54.1 & 49.0   \\
~+ reply in UL (UL) & 41.0 & 99.5 & 97.7 &  100 & 99.0 & 98.9 &  18.5 & 56.1 & 52.1 &  28.9 & 54.0 & 49.3   \\
~+ reply in UL + NE in UL (UL) & 28.8 & 99.5 & 98.7 &  100 & 99.8 & 99.1 & 17.6 & 55.9 & 51.2 & 29.6 & 55.1 & 49.4   \\
\bottomrule
\end{tabular}
\caption{Comparison of system prompts, for two generator models and in three retrieval settings: no retrieval, retrieval from English Wikipedia and from Wikipedia in user languages. Retrieval and reranking with BGE-m3. \newline
\textit{Main conclusions}: both models sometimes reply in English instead of the user language and it gets maximally addressed by explicitly specifying an instruction to generate response in the user language and translating the system prompt into the user language.}
\label{tab:prompts}
\end{table*}

\paragraph{Evaluation.}
Both MKQA and XOR-TyDi QA contain mostly short answer labels, e.g. a person name, a date etc. Following common RAG evaluation practice and~\citet{cora_asai}, we use lexical matching metrics, i.e. whether ground-truth or its tokens are contained in the generated answer. One key difference with \citep{cora_asai} is that we generate answers with off-the-shelf LLMs in a zero-shot setting, which tend to produce verbose answers, mostly consisting of full sentences rather than single-phrase outputs. While this is not a weakness, it requires adjusting metrics for reliable evaluation, e.g. prioritize \textit{recall} over precision and measure which percentage of tokens contained in the ground-truth label are contained in the response generated by the model.

In our preliminary experiments we noticed a pattern arising sometimes in the scenario with cross-lingual retrieval, when models generate a transliteration of named entities in other languages different from the one contained in the ground-truth label. This is again not a weakness of the system, but needs to be accounted in the evaluation metric. Since word-level matching fails to capture similarity in the described case, we propose to evaluate \textit{recall on character n-gram level}. We  first split ground-truth labels into tokens, extract all character 3-grams from each token and evaluate which percentage of such ngrams is present in the model-generated response, see Table~\ref{tab:charrecall} for illustration.

In addition to the task metric, we also control the correct language rate, CLR, which measures which percentage of model outputs are written in the user language. We detect languages using \verb|fasttext| library~\citep{fasttext1, fasttext2} and its \verb|lid.176.bin| model\footnote{\url{https://fasttext.cc/docs/en/language-identification.html}}. Due to high erroneous level of language identification for short sequences, we only evaluate the CRL metric for model responses longer than 20 characters.

\begin{table}[t!]
    \centering
    \footnotesize
    \begin{tabular}{p{2cm}|p{0.28cm}p{0.28cm}p{0.28cm}p{0.34cm}|p{0.3cm}p{0.3cm}p{0.3cm}p{0.3cm}}
    \toprule
    & \multicolumn{4}{c|}{Correct lang. rate} & \multicolumn{4}{c}{Char 3-gram recall} \\ \midrule
     & ko & fr & ru & en & ko & fr & ru & en \\
    \midrule
    \multicolumn{9}{c}{No retrieval} \\ \midrule 
    Llama-2-7B & 50.2 & 95.6 & 63.7 & 100 & 7.6 & 37.9 & 18.4 & 48.0 \\
    Solar-10.7B & 61.9 & 98.6 & 98.2 & 100 & 11.2 & 50.8 & 33.6 & 61.7 \\
    Mixtral-8x7B & 85.2 & 97.5 & 73.1 & 100 & 13.4 & 61.8 & 41.4 & 67.8 \\
    CommandR-35B & 99.6 & 97.4 & 98.3 & 100 & 18.6 & 52.6 & 36.2 & 58.4 \\ \midrule
    \multicolumn{9}{c}{Retrieval in English} \\ \midrule 
    Llama-2-7B & 4.3 & 62.8 & 0.8 & 100 & 17.4 & 58.9 & 21.1 & 70.8 \\
    Solar-10.7B & 53.1 & 99.7 & 99.7 & 100 & 18.4 & 64.5 & 56.7 & 74.5 \\
    Mixtral-8x7B & 89.0 & 95.7 & 34.4 & 100 & 22.7 & 64.8 & 32.9 & 73.3 \\
    CommandR-35B & 100 & 99.5 & 97.8 & 100 & 33.9 & 66.5 & 54.9 & 70.2 \\ \midrule 
    \multicolumn{9}{c}{Retrieval in user languages} \\ \midrule 
    Llama-2-7B & 7.3 & 47.6 & 5.1 & --- & 13.0 & 52.5 & 20.8 & --- \\
Solar-10.7B & 28.8 & 99.5 & 98.7 & --- & 17.6 & 55.9 & 51.2 & --- \\
Mixtral-8x7B & 92.5 & 97.1 & 64.4 & --- & 24.1 & 57.3 & 43.2 & --- \\
CommandR-35B & 100 & 99.8 & 99.1 & --- & 29.6 & 55.1 & 49.4 & --- \\
    \bottomrule
\end{tabular}
    \caption{Comparison of generator models (all models: instruction-tuned / aligned versions). Retrieval and reranking with BGE-m3. Prompt: "Reply short in UL + NE in UL (UL)" for non-English and "Reply short" for English. Llama-7B and Solar-10.7B are English-centric, while Mixtral-8x7B and Command-R-35B are multilingual by design.
    \newline \textit{Main conclusion:} using a multilingual-by-design model is essential to enable generation in a broad set of languages.
    % which prompt? 
    }
    \label{tab:models}
\end{table}

\section{Experimental details}
\paragraph{Retrieval.} We follow~\citet{cora_asai} and \citep{karpukhin-etal-2020-dense} and construct passages by splitting Wikipedia article into chunks of 100 words (or 100 Unicode characters for non whitespace separated languages, namely Chinese, Japanese, and Thai) and prepending the article title to each chunk. In most of the experiments we retrieve either from English Wikipedia (KILT version\footnote{\url{https://huggingface.co/datasets/facebook/kilt_wikipedia}}) or Wikipedia in the user language\footnote{\url{https://huggingface.co/datasets/wikimedia/wikipedia}}, but we also experiment with retrieving from concatenation of two mentioned Wikipedias and from Wikipedia in all considered languages. For each question in the evaluation data, we retrieve 50 relevant passages and pass them to the reranker to select top-5 relevant ones which will be inserted in the LLM context during generation.

\paragraph{Generation.} We use greedy decoding, limit generation to maximum 128 new tokens and run all experiments with model quantized into \verb|int4|.

\begin{table}[]
    \centering
    \footnotesize
    \begin{tabular}{p{1.75cm}|p{0.3cm}p{0.3cm}p{0.3cm}p{0.5cm}|p{0.3cm}p{0.3cm}p{0.3cm}p{0.4cm}}
    \toprule
    & \multicolumn{4}{c|}{Retrieval recall@5} & \multicolumn{4}{c}{Char 3-gram recall} \\ \midrule
     &  ko & fr & ru & en &  ko & fr & ru & en \\
    \midrule
    No retrieval &---  & --- & --- & --- & 18.6 & 52.6 & 36.2 & 58.4 \\  
    BGE-m3 & 61.5 & 78.4 & 77.1 & 88.5 & 33.9 & 66.5 & 54.9 & 70.2 \\
    SPLADE + QT & 60.9 & 72.0 & 71.9 & 78.5 & 32.9 & 63.6 & 51.3 & 66.0 \\
    BGE-m3 + QT & 61.5 & 78.4 & 77.1 & --- & 33.9 & 66.5 & 55.7 & --- \\
    Oracle & 100 & 100 & 100 & 100 & 44.1 & 70.4 & 60.5 & 71.2 \\  
    \bottomrule
    \end{tabular}
    \caption{Comparison of retrieval options (retrieval in English). Generator: Command-R-35B. BGE-m3: both retriever and reranker. SPLADE is coupled with MiniLM reranker. QT: query translation. SPLADE+QT for English means simply using SPLADE without QT. Recall@5 is reported for retrieval (before reranking). 
    \newline \textit{Main conclusion:} BGE-m3 enables reliable retrieval in the cross-lingual scenario. 
    }
    \label{tab:retrieval}
\end{table}

\paragraph{Evaluation.} We rely on the commonly-used SQUAD evaluation script\footnote{\url{https://github.com/allenai/bi-att-flow/blob/master/squad/evaluate-v1.1.py}}, but use it on the character 3-gram level, as discussed in Section~\ref{sec:main} and illustrated in Table~\ref{tab:charrecall}. We preprocess both ground-truth labels and predicted responses by lower-casing them, removing punctuation and articles.

\section{Results and discussion}
Table \ref{tab:1} summarizes the results across different languages on MKQA and XOR TyDi QA datasets. 
We observe a high performance improvement brought by RAG for all languages, but in many cases there is an important gap in performance in English and non-English.
In what follows we present multiple ablation studies to demonstrate steps needed to achieve shown results, to better understand the reasons behind the gap with English, and identify future research directions. We study the effect of the system prompt, generator model, retrieval system and language. We run ablations on three languages: French, Korean, and Russian.

\paragraph{Prompting strategy: importance of translating the system prompt into target languages and specifying the desired language of the response.}  Table \ref{tab:prompts} summarizes an impact of prompt formulation (defined in Table \ref{tab:prompts_demo}) on RAG performance  with English-centric \verb|SOLAR-10.7B| and multilingual \verb|Command-R-35B| models.

The left part reporting Correct Language Rate (CLR) allows us to assess how often the model replies in the user language.
Due to multilingual pretraining and instruction tuning, \verb|Command-R-35B|, equipped with the default system prompt ("Reply short (EN)"), replies in the user language in most, but not all, cases. Importantly, it gets "distracted" by the English context when retrieving from English Wikipedia and replies in English for around 50\% of non-English user queries. English-centric \verb|SOLAR-10.7B|, provided with the default system prompt, also often replies in English. These results demonstrate the need for using more advanced language-related prompting strategies for both models.

Explicitly specifying an instruction to reply in the given user language, %e.g. "Please reply in French", 
while keeping the system prompt itself in English ("\textit{+ reply in UL (EN)}"), substantially alleviates the problem of generation in English and correspondingly increases recall, but still does not enable correct language rate (CRL) close to 100\%. More generic prompt  with "meta-instruction" to reply in the same language  as the inout language (\textit{+ reply in same lang (EN)}) leads to considerably lower CRL than explicit language specification.

The further improvement in CRL (and thus recall) for both models is enabled by translating the system prompt into user languages. With the system prompt which includes explicit specification to generate in the given user language and is also written in the user language, both models achieve CRL > 95\% in most cases (except \verb|SOLAR-10.7B| for Korean). Such an approach is however less convenient in practice, as it requires language expertise to control the quality of translating prompts (see footnote 7) and dynamic selection of the system prompt based on the user query. 
\textbf{We believe that enabling multilingual LLMs to follow instructions within mixed-language prompts is an interesting research direction that would help to eliminate the need for the described ad-hoc prompting. }

The high CLR is necessary but not sufficient for high overall performance, as LLMs may use code-switching and tend to insert English named entities in their responses in user languages. 
We attempt to alleviate this issue by augmenting the system prompt with an explicit instruction to write all named entities in ULs. While it does slightly improve character 3-gram recall for Command-R in many cases, it does not solve the issue fully. \textbf{We believe that addressing the described code-switching problem is an important direction for future research}.

\paragraph{Generator model: importance of using a strong multilingual base model.}
Table~\ref{tab:models} compares four considered generator LLMs with and without retrieval. 
We find that  \verb|Command-R-35B| is the only model which consistently achieves high CLR and highest ranges of recall for all considered languages (with advanced prompts discussed above). Another considered multilingual-by-design model, \verb|Mixtral-8x7B|, reaches consistently high CLR and recall only for French which was present in its pretraining. 
English-centric \verb|LLAMA-2-chat-7B| most often replies in English.  Interestingly, English-centric \verb|SOLAR-10.7B| reaches high CLR 
and recall for French and Russian (with advanced prompts). This could be attributed to its strong capabilities in prompt understanding and accidental multilingual data present in pretraining. 

Despite \verb|Command-R-35B| being a leader model for non-English, its recall in English is much lower than of English-centric \verb|SOLAR-10.7B| which is possibly due to the "curse of multilinguality" effect. \textbf{This highlights the need for future models which would be fluent and accurate in both English and non-English.}

\paragraph{Retrieval: high performance of off-the-shelf multilingual retrievers in the in-domain setting.}
In our work we rely on a strong multilingual retriever and reranker, BGE-m3, which was shown by its authors to outperform other approaches on multilingual retrieval benchmarks. In Table~\ref{tab:retrieval} we evaluate its performance in the cross-lingual setting (documents in English and user queries in non-English), by comparing to the baselines involving query translation from user languages to English. We find that BGE-m3 outperforms a strong English model, SPLADE, used with translated queries. We note that BGE-m3 was trained on the datasets which also use Wikipedia as the document datastore, therefore in our experiments it is used in the in-domain setting. \textbf{The retrieval performance in the multilingual setting with domain-shift is yet to be explored.}

\begin{table}[]
    \centering
    \footnotesize
    \begin{tabular}{p{5.7cm}|ccc}
    \toprule
    \textbf{Error type} & \multicolumn{3}{c}{\textbf{Error count}} \\
    & \multicolumn{3}{c}{\textbf{(out of 50)}} \\
    & ru & zh & fr \\ \midrule 
    \multicolumn{4}{c}{System performance characteristics} \\ \midrule
    %All correct & 23\\
    Retrieved documents do not contain correct response & 4 & 9 & 8 \\
    Wrong response with correct retrieval & 4 & 7 & 3 \\
    Correct response with named entities in English & 5 & 6 & 0\\
    Correct response with different transliteration of named entities & 6 & 2 & 0 \\
    Correct response with code switching & 2 & 0 & 0\\
    Correct response with fluency issues & 1 & 1 & 0 \\
    Extra generated irrelevant text & 1 & 1 & 2 \\ \midrule
    \multicolumn{3}{c}{Data characteristics} \\ \midrule
    Ambiguous question (time-changing fact) & 7 & 8 & 5\\
    Ambiguous question (other) & 3 & 2 & 1 \\
    Typo in question & 1 & 0 & 0\\
    Fluency error in question & 1 & 0 & 1 \\
    Labels incomplete & 5 & 11 & 1\\
    Wrong labels & 1 & 4 & 7 \\
    Labels in English & 1 & 1 & 0\\
    \bottomrule
    \end{tabular}
    \caption{Statistics of manual inspection of 50 predictions for MKQA in Russian, Chinese, and French. Model: Command-R-35B. Retriever and reranker: BGE-m3, retrieval from English Wiki. Prompt: "Reply short in UL + NE in UL (UL)."
    }
    \label{tab:inspection}
\end{table}

\paragraph{Which language to retrieve from: highest performance with retrieving from multilingual Wikipedia.}
Table~\ref{tab:1} compares retrieval from English Wikipedia, Wikipedia in the user language, their union, and also Wikipedia in all considered languages. In the latter two cases with run retrieval over the embeddings of passages in multiple languages, so that the selected passages may be also in multiple languages. 

Comparing retrieval from English and user language, we observe different behavior on the two considered datasets. On the MKQA dataset, retrieval from English is more beneficial, which is expected since questions in MKQA were initially written by relying on the English Wikipedia and then translated into other languages. 
At the same time,  XOR-TyDi QA includes questions grounded in both English and user languages (see statistics in Table 2, \citealp{longpre-etal-2021-mkqa}), and we observe that retrieval from Wikipedia in the user language is more beneficial. 

Overall, we find that BGE-m3 also successfully manages to retrieve from the concatenated multilingual Wikipedia and thus dynamically choose the more appropriate datastore, often reaching performance higher than with any of the two monolingual Wikipedias. 

\paragraph{Best performing configuration to be used as a strong baseline.}
Based on the previous experiments, we highlight our best configuration, including \verb|Command-R-35B| generator, \verb|BGE-m3| retriever and reranker, the system prompt `Reply short in UL (UL)`, and retrieval from the concatenation of Wikipedia in various languages. 

\paragraph{Manual inspection of errors.}
To better analyze failure cases, we perform a manual analysis of predictions in French, Chinese, and Russian and report results in Table~\ref{tab:inspection}. We find that system improvements can be made at all steps, including retrieval, reading from the retrieved documents, addressing issues with code-switching and occasional fluency issues in non-English generation. Table~\ref{tab:retrieval} confirms gap in retrieval quality between English and non-English. Many examples are characterized by different transliteration of named entities which we take into account in evaluation, by computing lexical match metrics on the character n-gram level. \textbf{We underline that the possibility of various possible transliterations and code switching should be also kept in mind in the future development of evaluation metrics}. Finally, we notice several issues with evaluation data, including ambiguous questions and incomplete or wrong labels, as well as typos or fluency errors in questions.

\section{Conclusion}
In this work we study RAG in multilingual settings and build a strong pipeline to be used as a baseline in future works.  Better understanding of mRAG would enable reliable information access across different languages and cultures. We analyze an impact of each mRAG component impact on overall performance and provide guidelines and future research direction to further improve it.

Possible research directions include: 
\begin{itemize}%[noitemsep,topsep=0pt,parsep=0pt,partopsep=0pt]
\item \textit{The need for stronger multilingual LLMs and decoding strategies.} Our study highlights multilingual generation as a weakest part of the mRAG pipeline, especially with mixed-language context. We show that even strongest available multilingual LLMs can get distracted by the language of the prompt, and require ad-hoc prompting to enable consistent generation in the user language. Even then,  they are still prone to code-switching especially when writing named entities. We believe listed limitations could be addressed by including mixed-language examples in instruction tuning
or by developing specific decoding strategies.
\item \textit{LLM-based evaluation in multilingual settings.} In our work we rely on the lexical matching-based metrics due to their transparency and interpretability. At the same time, recent works use LLM-based evaluation which captures better semantic similarities but is currently underexplored in multilingual settings.
\item \textit{Multi-domain multilingual retrieval.} Current multilingual retrievers and rerankers are predominantly trained on Wikipedia-based data which could limit their applicability to other domains.
\end{itemize}

\section*{Limitations}

Following common practice in RAG and as a first step in mRAG, we run evaluation
on the open question answering task and with Wikipedia as the datastore. Important next steps include considering other tasks and domains.

Some of the standard practice in RAG which we left out of the scope of this study include query reformulation component and context post-processing (e.g. filtering irrelevant passages). These components are less relevant for the question answering datasets we studied, but will be more relevant for other tasks, and should be included in future work. 

We only considered single retriever and reranker model  \citep{chen2024bge} since this is the strongest open-source multilingual retrieval system available at the moment of our work, covering many different languages withing a single model.

\section*{Ethics Statement}
We do not anticipate negative societal impact from our work and on the reverse hope that it will help to broaden the accessibility of modern NLP to other languages.

\section*{Acknowledgments}
We gratefully appreciate the help of Shuai Wang, Inyoung Kim, Salah Ait  Mokhtar, Carlos Lassance, Beomseok Lee, 
 Tomi Silander, and Riccardo Volpi.

{
    \small
    \bibliographystyle{apalike}
    \bibliography{main}
}

\clearpage
\appendix
\section{Additional related works}
\label{app.related_works}
\lipsum[1-5]

\end{document}